\documentclass{article}

\usepackage{times}
\usepackage{graphicx} % more modern
\usepackage{subfigure} 

\usepackage{natbib}

\usepackage{algorithm}
\usepackage{algorithmic}

\usepackage[T1]{fontenc}
\usepackage{amsmath}
\usepackage{amssymb}

\usepackage{hyperref}

\newcommand{\pder}[1]{\nabla_{#1}}
\newcommand{\ddTheta}{\pder{\theta}}
\newcommand{\ddPhi}{\pder{\phi}}
\newcommand{\ddPhii}{\pder{\phi_i}}
\newcommand{\ddPsi}{\pder{\psi}}
\newcommand{\eqn}[1]{Eq.~\ref{#1}}
\newcommand{\sect}[1]{Section~\ref{#1}}
\newcommand{\tbl}[1]{Table~\ref{#1}}
\newcommand{\fig}[1]{Figure~\ref{#1}}
\newcommand{\wrt}{w.r.t.~}

\newcommand{\varL}{{\cal L}}

\usepackage[accepted]{icml2014}

\begin{document} 

\twocolumn[
\icmltitle{Neural Variational Inference and Learning in Belief Networks}

% It is OKAY to include author information, even for blind
% submissions: the style file will automatically remove it for you
% unless you've provided the [accepted] option to the icml2014
% package.
\icmlauthor{Andriy Mnih}{amnih@google.com}
\icmlauthor{Karol Gregor}{karolg@google.com}
\icmladdress{Google DeepMind}

% You may provide any keywords that you 
% find helpful for describing your paper; these are used to populate 
% the "keywords" metadata in the PDF but will not be shown in the document
\icmlkeywords{variational inference, belief networks, deep learning}

\vskip 0.3in
]

\begin{abstract}
Highly expressive directed latent variable models, such as sigmoid belief
networks, are difficult to train on large datasets because exact inference in
them is intractable and none of the approximate inference methods that have been
applied to them scale well.
We propose a fast non-iterative approximate inference method that uses a
feedforward network to implement efficient exact sampling from the variational
posterior. The model and this inference network are trained jointly by
maximizing a variational lower bound on the log-likelihood. Although the
naive estimator of the inference network gradient is too high-variance to be useful,
we make it practical by applying several straightforward model-independent variance
reduction techniques. Applying our approach to training sigmoid belief
networks and deep autoregressive networks, we show that it outperforms the
wake-sleep algorithm on MNIST and achieves state-of-the-art results on the
Reuters RCV1 document dataset.
\end{abstract}

\section{Introduction}

Compared to powerful globally-normalized latent variable models, such as deep
belief networks \citep{HOT} and deep Boltzmann machines \citep{salakhutdinov2009deep}, which can
now be trained on fairly large datasets, their purely directed counterparts
have been left behind due to the lack of efficient learning algorithms.  This
is unfortunate, because their modularity and ability to generate observations
efficiently make them better suited for integration into larger systems.

Training highly expressive directed latent variable models on large datasets is
a challenging problem due to the difficulties posed by inference. Although the
generality of Markov Chain Monte Carlo (MCMC) methods makes them
straightforward to apply to models of this type \citep{neal1992connectionist},
they tend to suffer from slow mixing and are usually too computationally
expensive to be practical in all but the simplest models. Such methods are also
difficult to scale to large datasets because they need to store the current state
of the latent variables for all the training observations between parameter
updates.

Variational methods \citep{jordan99variational} provide an optimization-based
alternative to the sampling-based Monte Carlo methods, and tend to be more
efficient. They involve approximating the exact posterior using a distribution
from a more tractable family, often a fully factored one, by maximizing a
variational lower bound on the log-likelihood \wrt the parameters of the
distribution.  For a small class of models, using such variational posteriors
allows the expectations that specify the parameter updates to be computed
analytically.  However, for highly expressive models such as the ones we are
interested in, these expectations are intractable even with the simplest
variational posteriors. This difficulty is usually dealt with by lower bounding
the intractable expectations with tractable one by introducing more variational
parameters, as was done for sigmoid belief nets by \citet{Saul96meanfield}.
However, this technique increases the gap between the bound being optimized and
the log-likelihood, potentially resulting in a poorer fit to the data.
In general, variational methods tend to be more model-dependent than
sampling-based methods, often requiring non-trivial model-specific derivations.

We propose a new approach to training directed graphical models that combines
the advantages of the sampling-based and variational methods. Its central
idea is using a feedforward network to implement efficient exact sampling from
the variational posterior for the given observation. We train this inference
network jointly with the model by maximizing the variational lower bound on the
log-likelihood, estimating all the required gradients using samples from the
inference network. Although naive estimate of the gradient for the inference
network parameters is unusable due to its high variance, we make the approach
practical by applying several straightforward and general variance reduction techniques.
The resulting training procedure for the inference network can be seen as an
instance of the REINFORCE algorithm \citep{williams1992simple}.  
Due to our use of stochastic feedforward networks for performing inference we
call our approach Neural Variational Inference and Learning (NVIL).

Compared to MCMC methods, where many iterations over the latent variables
are required to generate a sample from the exact posterior and successive
samples tend to be highly correlated, NVIL does not suffer from mixing
issues as each forward pass through the inference network generates an
independent exact sample from the variational posterior. In addition to being
much faster than MCMC, our approach has the additional advantage of not needing
to store the latent variables for each observation and thus is not only more
memory efficient but also applicable to the pure online learning setting,
where each training case is seen once before being discarded.

In contrast to other work on scaling up variational inference, NVIL can handle both discrete
and continuous latent variables (unlike \citet{kingma2013auto,rezende2014stochastic}) as well variational posteriors with
complex dependency structures (unlike \citet{ranganath2013black}). Moreover, the variance reduction methods we
employ are simple and model-independent, unlike the more sophisticated
model-specific control variates of \citet{conf/icml/PaisleyBJ12}.

Though the idea of training an inference model by following the gradient of the
variational bound has been considered before, it was dismissed as
infeasible \citep{dayan1996varieties}. Our primary contribution is to show how to reduce the
variance of the naive gradient estimator to make it practical without narrowing
its range of applicability. We also show that the resulting method
trains sigmoid belief networks better than the wake-sleep algorithm
\citep{HintonWakeSleep}, which is the only algorithm we are aware of that is capable of
training the same range of models efficiently.  Finally, we demonstrate the
effectiveness and scalability of NVIL by using it to achieve state-of-the-art results
on the Reuters RCV1 document dataset.

\section{Neural variational inference and learning}

\subsection{Variational objective}

Suppose we are interested in training a latent variable model $P_{\theta}(x,h)$
with parameters $\theta$. We assume that exact inference in the model is
intractable and thus maximum likelihood learning is not an option.  For
simplicity, we will also assume that all the latent variables in the model are
discrete, though essentially the same approach applies if some or all of the
variables are continuous.  

We will train the model by maximizing a variational lower bound on the marginal
log-likelihood. Following the standard variational inference approach
\citep{jordan99variational}, given an observation $x$, we introduce a distribution
$Q_{\phi}(h|x)$ with parameters $\phi$, which will serve as an approximation to
its exact posterior $P_{\theta}(h|x)$. The variational posterior $Q$ will have
a simpler form than the exact posterior and thus will be easier to work with.

The contribution of $x$ to the log-likelihood can then be lower-bounded as
follows \citep{jordan99variational}:
\begin{align}
    \log P_{\theta}(x) & =   \log \sum_h P_{\theta}(x,h) \nonumber \\
                     & \ge \sum_h Q_{\phi}(h|x) \log \frac{P_{\theta}(x,h)}{Q_{\phi}(h|x)} \nonumber \\
                     & = E_Q[\log P_{\theta}(x,h) - \log Q_{\phi}(h|x)] \\
\label{eqn:bound}
                     & = \varL(x, \theta, \phi) \nonumber.
\end{align}
By rewriting the bound as 
\begin{align}
    \varL(x, \theta, \phi) = \log P_{\theta}(x) - KL(Q_{\phi}(h|x), P_{\theta}(h|x)),
\end{align}
we see that its tightness is determined by the Kullback-Leibler (KL) divergence
between the variational distribution and the exact posterior. Maximizing the
bound with respect to the parameters $\phi$ of the variational distribution
makes the distribution a better approximation to the posterior (\wrt the
KL-divergence) and tightens the bound. 

In contrast to most applications of variational inference where the variational
posterior for each observation is defined using its own set of variational
parameters, our approach does not use any local variational parameters.
Instead, we use a flexible feedforward model to compute the variational
distribution from the observation. We call the model mapping $x$ to
$Q_{\phi}(h|x)$ the \textit{inference network}. The architecture of the
inference network is constrained only by the requirement that $Q_{\phi}(h|x)$ it
defines has to be efficient to evaluate and sample from.
Using samples from the inference network we will be able to compute gradient
estimates for the model and inference network parameters for a large class of
highly expressive architectures, without having to deal with
architecture-specific approximations.

Given a training set ${{\cal D}}$, consisting of observations $x_1, ..., x_D$, we train
the model by (locally) maximizing $\varL({{\cal D}}, \theta, \phi) = \sum_i \varL(x_i, \theta,
\phi)$ using gradient ascent \wrt to the model and inference network
parameters. To ensure scalability to large datasets, we will perform stochastic
optimization by estimating gradients on small minibatches of randomly sampled
training cases.

\subsection{Parameter gradients}

The gradient of the variational bound for a single observation $x$ \wrt to the
model parameters is straightforward to derive and has the form
\begin{align}
        \ddTheta \varL(x) & = E_Q \left[\ddTheta \log P_{\theta}(x,h) \right],
\end{align}
where we left $\theta$ and $\phi$ off the list of the arguments of $\varL$ to simplify the notation.
The corresponding gradient \wrt to the inference network parameters is somewhat more involved:
\begin{align}
\label{eqn:NaiveGrads}
    \ddPhi \varL(x) = E_Q [&( \log P_{\theta}(x,h) - \log Q_{\phi}(h|x) ) \nonumber \\
                       & \times \ddPhi \log Q_{\phi}(h|x)],
\end{align}
We give its derivation in the supplementary material.

As both gradients involve expectations which are intractable in all but a
handful of special cases, we will estimate them with Monte Carlo integration,
using samples from the inference network. Having generated $n$ samples $h^{(1)}, ..., h^{(n)}$ from
$Q_{\phi}(h|x)$, we compute
\begin{align}
\label{eqn:ModelGradEstimator}
    \ddTheta \varL(x) \approx \frac{1}{n} \sum_{i=1}^n \ddTheta \log P_{\theta}(x,h^{(i)})
\end{align}
and 
\begin{align}
\label{eqn:NaiveGradEstimator}
    \ddPhi \varL(x) \approx \frac{1}{n} \sum_{i=1}^n & ( \log P_{\theta}(x,h^{(i)}) - \log Q_{\phi}(h^{(i)}|x) ) \nonumber \\
                       & \times \ddPhi \log Q_{\phi}(h^{(i)}|x).
\end{align}
The above gradient estimators are unbiased and thus can be used to perform
stochastic maximization of the variational objective using a suitable learning
rate annealing schedule. The speed of convergence of this procedure,
however, depends heavily on the variance of the estimators used, as we will
see in \sect{sec:Experiments}.

The model gradient estimator (\ref{eqn:ModelGradEstimator}) is well-behaved and
does not pose a problem.
 The variance of the
inference network gradient estimator (\ref{eqn:NaiveGradEstimator}), however, can be very high due to the scaling
of the gradient inside the expectation by a potentially large term.
As a result, learning variational parameters with updates based on this
estimator can be unacceptably slow.  In fact, it is widely believed that
learning variational parameters using gradient estimators of the form
(\ref{eqn:NaiveGradEstimator}) is infeasible
\citep{hinton1994autoencoders,dayan1996varieties,kingma2013auto}.
In the next section we will show how to make this approach practical by applying
variance reduction techniques.

\subsection{Variance reduction techniques}
\label{sec:variance_reduction}

Though gradient estimates computed using \eqn{eqn:NaiveGradEstimator} are
usually too noisy to be useful in practice, it is easy to reduce their variance to
a manageable level with the following model-independent techniques.

\subsubsection{Centering the learning signal}
\label{sec:centering}

Inspecting \eqn{eqn:NaiveGrads}, we see that we are using
\begin{align}
\label{eqn:learningSignal}
    l_{\phi}(x,h) = \log P_{\theta}(x,h) - \log Q_{\phi}(h|x)
\end{align}
as the learning signal for the inference
network parameters, and thus are effectively fitting $\log Q_{\phi}(h|x)$ to
$\log P_{\theta}(x,h)$.  This might seem surprising, given that we want the
inference network $Q_{\phi}(h|x)$ to approximate the posterior distribution
$P_{\theta}(x|h)$, as opposed to the joint distribution $P_{\theta}(x,h)$. It
turns out however that using the joint instead of the posterior distribution in
\eqn{eqn:NaiveGrads} does not affect the value of the expectation. To see that
we start by noting that
\begin{align}
E_Q [ \ddPhi \log Q_{\phi}(h|x)] & = E_Q \left[ \frac{\ddPhi  Q_{\phi}(h|x)}{Q_{\phi}(h|x)} \right] \nonumber \\ 
      & = \ddPhi E_Q [1] = 0.
\end{align}
Therefore we can subtract any $c$ that does not depend on $h$ from the learning signal in
\eqn{eqn:NaiveGrads} without affecting the value of the expectation:
\begin{align}
\label{eqn:baselineIdentity}
& E_Q [(l_{\phi}(x,h) - c) \ddPhi \log Q_{\phi}(h|x)] \nonumber \\
& = E_Q [l_{\phi}(x,h) \ddPhi \log Q_{\phi}(h|x)] - c E_Q [\ddPhi \log Q_{\phi}(h|x)] \nonumber \\
& = E_Q [l_{\phi}(x,h) \ddPhi \log Q_{\phi}(h|x)].
\end{align}
And as $\log P_{\theta}(x,h) = \log P_{\theta}(h|x) + \log P_{\theta}(x)$
and $\log P_{\theta}(x)$ does not depend on $h$, using $P_{\theta}(h|x)$
in \eqn{eqn:NaiveGrads} in place of $P_{\theta}(x,h)$ does not affect the value of the expectation.

This equivalence allows us to compute the learning signal efficiently, without
having to evaluate the intractable $P_{\theta}(h|x)$ term. The price we pay for
this tractability is the much higher variance of the estimates computed using
\eqn{eqn:NaiveGradEstimator}. Fortunately, \eqn{eqn:baselineIdentity} 
suggests that we can reduce the variance by subtracting a carefully chosen $c$
from the learning signal. The simplest option is to make $c$ a parameter and
adapt it as learning progresses.  However, $c$ will not be able capture the
systematic differences in the learning signal for different observations $x$,
which arise in part due to the presence of the $\log P_{\theta}(x)$ term. 
Thus we can reduce the gradient variance further by subtracting an
observation-dependent term $C_{\psi}(x)$ to minimize those differences. Doing this
does not affect the expected value of the gradient estimator because 
$C_{\psi}(x)$ does not depend on the latent variables. Borrowing a name from the
reinforcement learning literature we will refer to $c$ and $C_{\psi}(x)$ as
\textit{baselines}. We will elaborate on this connection in
\sect{sec:reinforce}.

We implement the input-dependent baseline $C_{\psi}(x)$  using a neural network and train it
to minimize the expected square of the centered learning signal $E_Q[(l_{\phi}(x,h) -
C_{\psi}(x) - c)^2]$. Though this approach to fitting the baseline does not
result in the maximal variance reduction, it is simpler and in our experience
works as well as the optimal approach of \citet{Weaver01theoptimal} which requires
taking into account the magnitude of the gradient of the inference network
parameters. We also experimented with per-parameter baselines but found
that they did not improve on the global ones. Finally, we 
note that incorporating baselines into the learning signal can be seen as 
using simple control variates. In contrast
to the more elaborate control variates (e.g. of
\citet{conf/icml/PaisleyBJ12}), baselines do not depend on the form of the
model or of the variational distribution and thus are easier to use.

\subsubsection{Variance normalization}

Even after centering, using $l_{\phi}(x,h)$ as the learning signal is
non-trivial as its average magnitude can change dramatically, and not
necessarily monotonically, as training progresses. This variability makes
training an inference network using a fixed learning rate difficult.  We
address this issue by dividing the centered learning signal by a running
estimate of its standard deviation. This normalization ensures that the signal
is approximately unit variance, and can be seen as a simple and efficient
way of adapting the learning rate. To ensure that we stop learning when the
magnitude of the signal approaches zero, we apply variance normalization only
when the estimate of the standard deviation is greater than 1. The algorithm
for computing NVIL parameter updates using the variance reduction techniques
described so far is provided in the supplementary material.

\subsubsection{Local learning signals}
\label{sec:layer_signals}

So far we made no assumptions about the structure of the model or the inference
network. However, by taking advantage of their conditional independence
properties we can train the inference network using simpler and less noisy
local learning signals instead of the monolithic global learning signal
$l_{\phi}(x,h)$.  Our approach to deriving a local signal for a set of
parameters involves removing all the terms from the global signal that do not
affect the value of the resulting gradient estimator.

We will derive the layer-specific learning signals for the common case of both
the model and the inference network having $n$ layers of latent variables.  The
model and the variational posterior distributions then naturally factor as
\begin{align}
    P_{\theta}(x,h) = & P_{\theta}(x|h^1) \prod \nolimits_{i=1}^{n-1} P_{\theta}(h^i|h^{i+1}) P_{\theta}(h^n), \label{eqn:layeredModel} \\
    Q_{\phi}(h|x) = & Q_{\phi^1}(h^1|x) \prod \nolimits_{i=1}^{n-1} Q_{\phi^{i+1}}(h^{i+1}|h^i) \label{eqn:layeredInfNet},
\end{align}
where $h^i$ denotes the latent variables in the $i^{th}$ layer and $\phi^i$ the
parameters of the variational distribution for that layer.  We will also use
$h^{i:j}$ to denote the latent variables in layers $i$ through $j$.

To learn the parameters of the the variational distribution for layer $i$ ,
we need to compute the following gradient:
\begin{align*}
    \ddPhii \varL(x) = E_{Q(h|x)} [& l_{\phi}(x,h) \ddPhii \log Q_{\phi_i}(h^i|h^{i-1})].
\end{align*}
Using the law of iterated expectation we can rewrite the expectation \wrt
$Q(h|x)$ as
\begin{align*}
    \ddPhii & \varL(x) =  E_{Q(h^{1:i-1}|x)}[ \\ 
    & E_{Q(h^{i:n}|h^{i-1})}[ l_{\phi}(x,h) \ddPhii \log Q_{\phi_i}(h^i|h^{i-1})]|h^{i-1}]],
\end{align*}
where we also used the fact that under the variational posterior, $h^{i:n}$ is
independent of $h^{1:i-2}$ and $x$, given $h^{i-1}$.
As a consequence of \eqn{eqn:baselineIdentity}, when computing the expectation
\wrt $Q(h^{i:n}|h^{i-1})$, all the terms in the learning signal that do not
depend on $h^{i:n}$ can be safely dropped without affecting the result. 
This gives us the following local learning signal for layer $i$:
\begin{align}
\label{eqn:localSignal}
l^i_{\phi}(x,h) = \log P_{\theta}(h^{i-1:n}) - \log Q_{\phi}(h^{i:n}|h^{i-1}).
\end{align}
To get the signal for the first hidden layer we simply use $x$ in place of $h^0$,
in which case we simply recover the global learning signal. For hidden layers
$i>1$, however, the local signal involves fewer terms than $l_{\phi}(x,h)$ and
thus can be expected to be less noisy.
As we do not assume any within-layer structure, \eqn{eqn:localSignal}
applies to models and inference networks whether or not $Q_{\phi}(h^i|h^{i-1})$
and $P_{\theta}(h^i|h^{i+1})$ are factorial.

Since local signals can be significantly different from each other, we use
separate baselines and variance estimates for each signal. For layers $i>1$,
the input-dependent baseline $C_{\psi}(x)$ is replaced by $C^i_{\psi_i}(h^{i-1})$.

In some cases, further simplification of the learning signal is possible, yielding a
different signal per latent variable. We leave exploring this as future work.

\section{Related work}

\subsection{Feedforward approximations to inference}

The idea of training an approximate inference network by optimizing a
variational lower bound is not new. It goes back at least to
\citet{hinton1994autoencoders}, who derived the variational objective from the
Minimum Description Length (MDL) perspective and used it to train linear
autoencoders. Their probabilistic encoder and decoder correspond to our
inference network and model respectively.  However, they computed the gradients
analytically, which was possible due to the simplicity of their model, and
dismissed the sampling-based approach as infeasible due to noise.

\citet{salakhutdinov2010efficient} proposed using a feedforward ``recognition''
model to perform efficient input-dependent initialization for the mean field
inference algorithm in deep Boltzmann machines. As the recognition model is
trained to match the marginal probabilities produced by mean field inference,
it inherits the limitations of the inference procedure, such as the
inability to model structured posteriors. In contrast, in NVIL the inference
net is trained to match the true posterior directly, without involving an
approximate inference algorithm, and thus the accuracy of the fit is
limited only by the expressiveness of the inference network itself.

Recently a method for training nonlinear models with
continuous latent variables, called Stochastic Gradient Variational
Bayes (SGVB), has been proposed by \citet{kingma2013auto} and \citet{rezende2014stochastic}.
Like NVIL, it involves using feedforward models to perform approximate
inference and trains them by optimizing a sampling-based estimate of the
variational bound on the log-likelihood. However, SGVB is considerably less
general than NVIL, because it uses a gradient estimator obtained by taking
advantage of special properties of real-valued random variables and thus is not
applicable to models with discrete random variables. Moreover, unlike NVIL,
SGVB method cannot handle inference networks with nonlinear dependencies
between latent variables. The ideas of the two methods are complementary however, and
NVIL is likely to benefit from the SGVB-style treatment of continuous-valued
variables, while SGVB might converge faster using the variance reduction
techniques we proposed.

\citet{gregor2013deep} have recently proposed a related algorithm for training
sigmoid belief network like models based on the MDL framework.
They also use a feedforward model to perform approximate
inference, but concentrate on the case of a deterministic inference network and
can handle only binary latent variables. The inference network is trained by
backpropagating through binary thresholding units, ignoring the thresholding
nonlinearities, to approximately minimize the coding cost of the joint
latent-visible configurations. This approach can be seen as approximately maximizing a
looser variational lower bound than (\ref{eqn:bound}) due to the absence of the
entropy term.

An inference network for efficient generation of samples from the approximate
posterior can also be seen as a probabilistic generalization of the approximate
feedforward inference methods developed for sparse coding models in the last
few years \citep{koray-psd-08,bradley2008differential,gregor-icml-10}.

\subsection{Sampling-based variational inference}

Like NVIL, Black Box Variational Inference \citep[BBVI,][]{ranganath2013black} learns the
variational parameters of the posterior by optimizing the variational bound
using sampling-based gradient estimates, which makes it applicable to a large
range of models. However, unlike NVIL, BBVI follows the traditional approach of
learning a separate set of variational parameters for each observation and does
not use an inference network. Moreover, BBVI uses a fully-factorized mean field
approximation to the posterior, which limits its power.

\subsection{The wake-sleep algorithm}

NVIL shares many similarities with the wake-sleep algorithm
\citep{HintonWakeSleep}, which enjoys the same scalability and applicability to
a wide range of models. This algorithm was introduced for training Helmholtz
machines \citep{dayan1995helmholtz}, which are multi-layer belief networks augmented
with recognition networks. These recognition networks are used for approximate inference
and are directly analogous to NVIL inference networks. Wake-sleep alternates between
updating the model parameters in the wake phase and the recognition network
parameters in the sleep phase. The model parameter update is based on the
samples generated from the recognition network on the training data and is
identical to the NVIL one (\eqn{eqn:ModelGradEstimator}). However, in contrast
to NVIL, the recognition network parameters are learned from samples generated
by the model. In other words, the recognition network is trained to recover the
hidden causes corresponding to the samples from the model distribution by following
the gradient
\begin{align}
\label{eqn:wakeSleepUpdate}
        \ddPhi \varL(x) & = E_{P_{\theta}(x,h)} \left[\ddPhi \log Q_{\phi}(h|x) \right].
\end{align}
Unfortunately, this update does not optimize the same objective as the model
parameter update, which means that the wake-sleep algorithm does not optimize a
well-defined objective function and is not guaranteed to converge. This is the
algorithm's main weakness, compared to NVIL, which optimizes a variational
lower bound on the log-likelihood.

The wake-sleep gradient for recognition network parameters does have the
advantage of being much easier to estimate than the corresponding gradient of
the variational bound. In fact, the idea of training the recognition networks
using the gradient of the bound was mentioned in \citep{hinton1994autoencoders} and \citep{dayan1996varieties} but not
seriously considered due concerns about the high variance of the
estimates. In \sect{sec:Experiments} we show that while the naive estimator of
the gradient given in \eqn{eqn:NaiveGradEstimator} does exhibit high variance,
the variance reduction techniques from \sect{sec:variance_reduction}
improve it dramatically and make it practical.

\subsection{REINFORCE}
\label{sec:reinforce}

Using the gradient (\ref{eqn:NaiveGrads}) to train the inference network can be
seen as an instance of the REINFORCE algorithm \citep{williams1992simple} from
reinforcement learning (RL), which adapts the parameters of a stochastic model
to maximize the external reward signal which depends on the model's output.
Given a model $P_{\theta}(x)$ and a reward signal $r(x)$, REINFORCE updates the
model parameters using the rule
\begin{align}
\label{eqn:reinforce}
\Delta \theta \propto E_P[(r(x) - b) \ddTheta \log P_{\theta}(x)].
\end{align}
We can view NVIL as an application of REINFORCE on the per-training-case
basis, with the inference network corresponding to
the stochastic model, latent state $h$ to the output, and the learning signal
$l_{\phi}(x,h)$ to the reward. The term $b$ in \eqn{eqn:reinforce}, called a
\textit{baseline} in the RL literature, is a hyperparameter that can be
adapted to reduce the variance of the parameter update. Thus it serves the same
function as $c$ and $C_{\psi}(x)$ that we subtract from the learning signal to
center it in \sect{sec:centering}. The considerable body of work on baselines
and other variance reduction methods done in the RL community \citep[e.g.][]{GreensmithBB04} 
is likely to contain additional techniques relevant for training
inference networks.

\section{Experimental results}

We performed two sets of experiments, with the first set intended to evaluate
the effectiveness of our variance reduction techniques and to compare NVIL's
performance to that of the wake-sleep algorithm. In the second set of
experiments, we demonstrate NVIL's ability to handle larger real-world datasets by
using it to train generative models of documents.

\subsection{Experimental protocol}

We trained all models using stochastic gradient ascent using minibatches of 20
observations sampled randomly from the training data. The gradient estimates
were computed using a single sample from the inference network.  For each
dataset, we created a validation set by removing a random subset of 100
observations from the training set. The only form of regularization we used was
early stopping based on the validation bound, implemented by keeping track of
the parameter configuration with the best validation score seen so far.  We
implemented each input-dependent baseline using a neural network with a single
hidden layer of 100 tanh units.

We used fixed learning rates because we found them to produce superior results
to the annealing schedules we experimented with. The learning rates we report
were selected based on the validation set performance in preliminary experiments
with smaller models. We always make the learning rate for inference network
five times smaller than for the model (which is the one we report), as we found
this to improve performance.  We used inference networks with layered structure
given by \eqn{eqn:layeredInfNet}, without dependencies within each layer except in the
experiment with autoregressive inference networks. All multi-layer inference
networks were trained using layer-specific learning signals from
\sect{sec:layer_signals}.

As the models we train are intractable, we cannot compute the exact
log-likelihoods for them. Instead we report the estimates of the variational
bound (\ref{eqn:bound}) computed using 10 samples from the inference network,
which we found to be sufficient to get the accurate bound estimates. We expect
this approach to underestimate the log-likelihood considerably, but leave
finding more direct and thus less pessimistic evaluation methods as future
work.

\subsection{Modelling images of digits}
\label{sec:Experiments}

\begin{figure*}%
\begin{minipage}{0.5\textwidth}
\includegraphics[width=.99\textwidth]{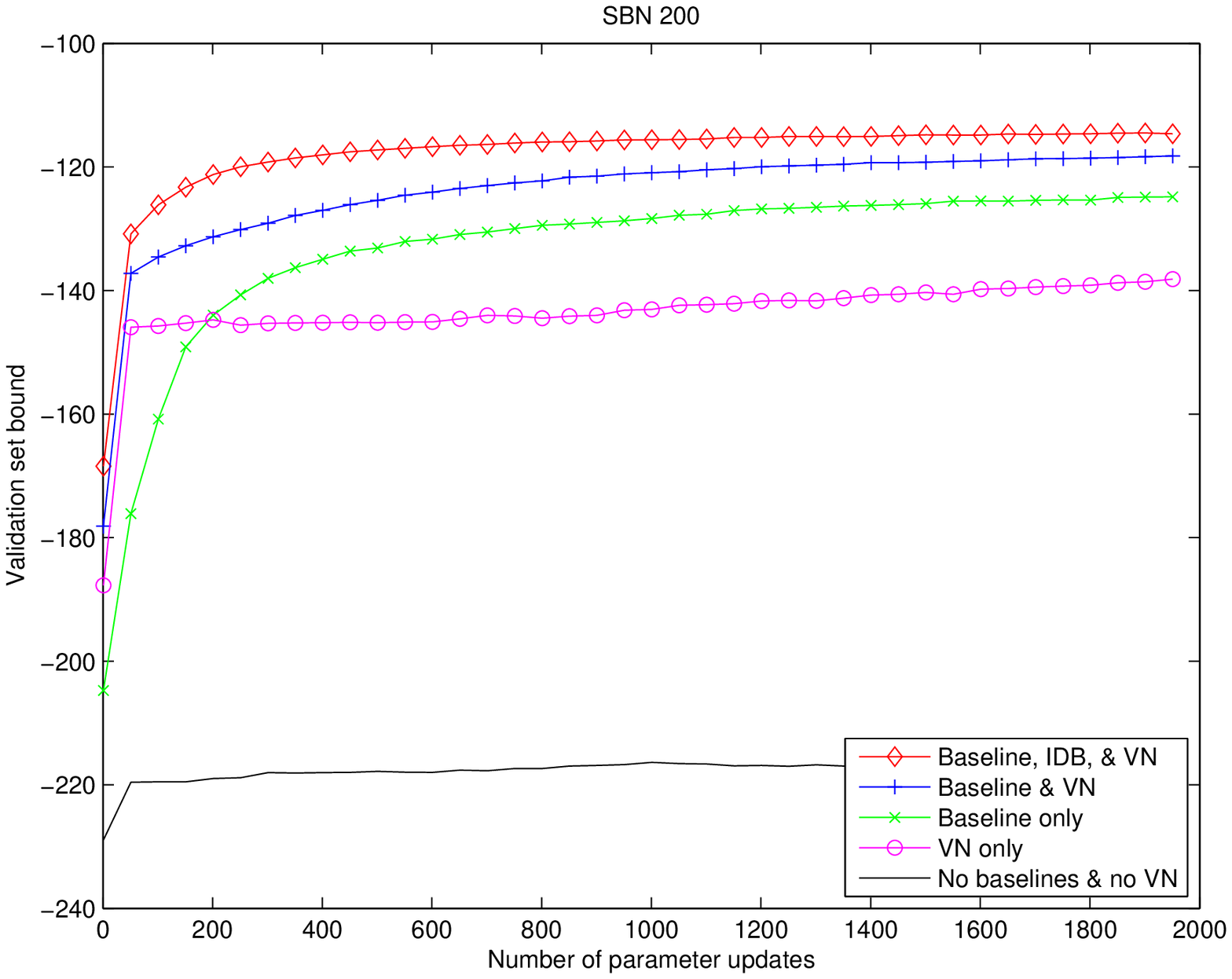}
\end{minipage}
\begin{minipage}{0.5\textwidth}
\includegraphics[width=.99\textwidth]{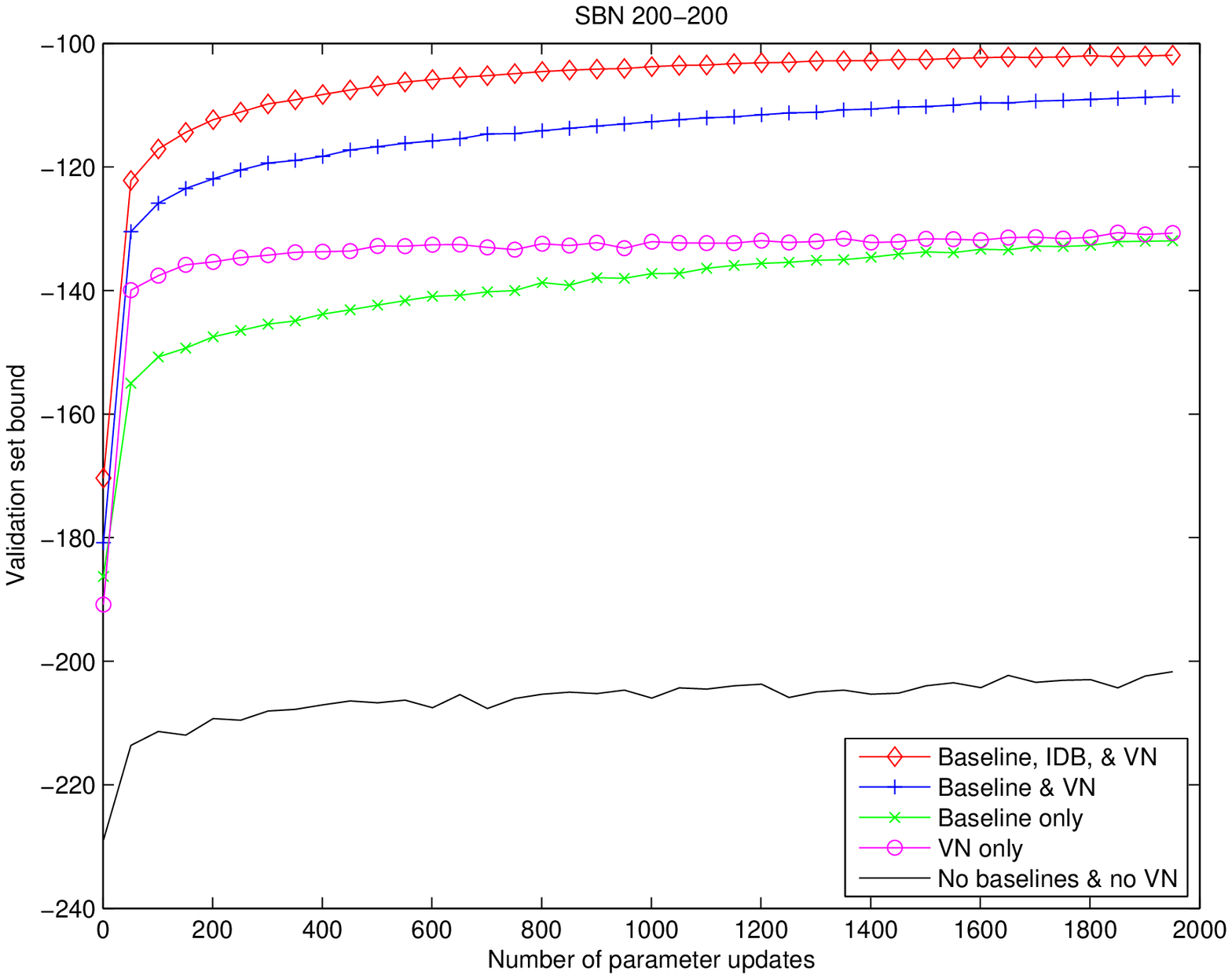}
\end{minipage}
\caption{Bounds on the validation set log-likelihood for an SBN with (Left) one 
and (Right) two layers of 200 latent variables.  Baseline and IDB refer to the
input-independent and the input-dependent baselines respectively. VN is variance
normalization.  }
\label{fig:mnist_bounds}
\end{figure*}

Our first set of experiments was performed on the binarized version of the
MNIST dataset, which has become the standard benchmark for
evaluating generative models of binary data. The dataset consists of 70,000
$28\times28$ binary images of handwritten digits, partitioned into a
60,000-image training set and 10,000-image test set. We used the binarization
of \citet{salakhutdinov2008}, which makes our scores directly comparable to
those in the literature.

We used $3\times 10^{-4}$ as the learning rate for training models with NVIL on
this dataset. Centering the input vectors by subtracting the mean vector was
essential for making the inference networks and input-dependent baselines 
work well.

To demonstrate the importance of variance reduction techniques, we trained two
SBNs using a range of variance control settings. The first SBN had a single
layer of 200 latent variables, while the second one had two layers of 200 variables
each.  \fig{fig:mnist_bounds} shows the
estimate of the variational objective on the validation set plotted against the number of
parameter updates. For both models, it is clear that using all three techniques -- the input-dependent
and input-independent baselines along with variance normalization -- is essential for
best performance.  However, of the three techniques, the input-dependent
baseline appears to be the least important. Comparing the plots for the two
models suggests that variance reduction becomes more important for larger
models, with the gap between the best combination and the others (excluding the
very worst one) widening. For both models, learning with all three variance
reduction techniques disabled makes barely any progress and is clearly infeasible.

We found that disabling layer-specific learning signals had little effect on
the performance of the resulting model. The difference was about 0.4 nats for an
SBN with two or three layers of latent variables.

\begin{table}[t]
\caption{Results on the binarized MNIST dataset. ``Dim'' is the number of latent
variables in each layer, starting with the deepest one. NVIL and WS refer to
the models trained with NVIL and wake-sleep respectively.  NLL is the negative
log-likelihood for the tractable models and an estimate of it for the
intractable ones.  
}
\label{tbl:MNISTResults}
\begin{center}
\begin{small}
\begin{sc}
\begin{tabular}{|l|r|r|r|r|}
\hline
Model              &  Dim          &  \multicolumn{2}{|c|}{Test NLL}  \\
                   &               &  \multicolumn{1}{|c|}{NVIL}   &  \multicolumn{1}{|c|}{WS}  \\
\hline
SBN                & 200           & 113.1 & 120.8 \\
SBN                & 500           & 112.8 & 121.4 \\
SBN                & 200-200       & 99.8  & 107.7 \\
SBN                & 200-200-200   & 96.7  & 102.2 \\
SBN                & 200-200-500   & 97.0  & 102.3 \\
fDARN              & 200           & 92.5  &  95.9 \\
fDARN              & 500           & 90.7  &  97.2 \\
\hline
fDARN               & 400           & \multicolumn{2}{|c|}{96.3} \\
DARN                & 400           & \multicolumn{2}{|c|}{93.0}  \\
NADE                & 500           & \multicolumn{2}{|c|}{88.9} \\
RBM (CD3)           & 500           & \multicolumn{2}{|c|}{105.5} \\
RBM (CD25)          & 500           & \multicolumn{2}{|c|}{86.3}  \\
MoB                 & 500           & \multicolumn{2}{|c|}{137.6} \\
\hline
\end{tabular}
\end{sc}
\end{small}
\end{center}
\vspace{-0.2in}
\end{table}

We next compared NVIL to the wake-sleep algorithm, which is its closest
competitor in terms of scalability and breadth of applicability, by training a
range of models using both algorithms. Wake-sleep training used a learning
rate of $1\times 10^{-4}$, as we found this algorithm to be more sensitive to
the choice of the learning rate than NVIL, performing considerably better with
lower learning rates.  The results, along with some baselines from the
literature, are shown in \tbl{tbl:MNISTResults}.  We report only the
means of the bound estimates as their standard deviations were all very small,
none exceeding 0.1 nat.  We can see that models trained with NVIL have
considerably better bounds on the log-likelihood, compared to their wake-sleep
counterparts, with the difference ranging from 3.4 to 8.6 nats. Additional
layers make SBNs perform better, independently of the training method.
Interestingly, single-layer fDARN \citep{gregor2013deep} models,
which have autoregressive connections between the latent variables,
 perform better than any of the SBN models
trained using the same algorithm. Comparing to results from the literature,
we see that all the SBN and fDARN models we trained perform much better than 
a mixture of 500 factorial Bernoulli distributions (MoB) but not as well as 
the deterministic Neural Autoregressive Distribution Estimator (NADE) \citep{larochelle2011}.
The NVIL-trained fDARN models with 200 and 500 latent
variables also outperform the fDARN (as well as the more expressive DARN)
model with 400 latent variables from \citep{gregor2013deep}, which were trained using an MDL-based algorithm.
The fDARN and multi-layer SBN models trained using NVIL also
outperform a 500-hidden-unit RBM trained with 3-step contrastive divergence
(CD), but not the one trained with 25-step CD \citep{salakhutdinov2008}. However, both sampling and
CD-25 training in an RBM is considerably more expensive 
than sampling or NVIL training for any of our models.

The sampling-based approach to computing gradients allows NVIL to handle
variational posteriors with complex dependencies. To demonstrate this ability,
we retrained several of the SBN models using inference networks with
autoregressive connections within each layer. These networks can capture the
dependencies between variables within layers and thus are considerably more
expressive than the ones with factorial layers. Results in \tbl{tbl:autoreg}
indicate that using inference networks with autoregressive connections produces better
models, with the single-layer models exhibiting large gains. 

\begin{table}[t]
\caption{The effect of using autoregressive connections in the inference
network.  ``Dim'' is the number of latent variables in each layer, starting with
the deepest one.  ``Test NLL'' is an estimate of the lower bound on the
log-likelihood on the MNIST test set. 
''Autoreg'' and ``Factorial'' refer to using inference networks with and without
autoregressive connections respectively.  }
\label{tbl:autoreg}
\vskip 0.15in
\begin{center}
\begin{small}
\begin{sc}
\begin{tabular}{|l|r|r|r|r|}
\hline
Model              &  Dim          &  \multicolumn{2}{|c|}{Test NLL}  \\
                   &               &  \multicolumn{1}{|c|}{Autoreg}   &  \multicolumn{1}{|c|}{Factorial}  \\
\hline
SBN                & 200           & 103.8 & 113.1 \\
SBN                & 500           & 104.4 & 112.8 \\
SBN                & 200-200-200   &  94.5 & 96.7  \\
SBN                & 200-200-500   &  96.0 & 97.0  \\
\hline
\end{tabular}
\end{sc}
\end{small}
\end{center}
\vskip -0.1in
\end{table}

\subsection{Document modelling}

We also applied NVIL to the more practical task of document modelling. The goal
is to train a generative model of documents which are represented as
vectors of word counts, also known as bags of words. We trained two simple
models on the 20 Newsgroups and Reuters Corpus Volume I (RCV1-v2) datasets,
which have been used to evaluate similar models in \citep{salakhutdinov2009replicated, larochelle2012neural}.
20 Newsgroups is a fairly small dataset of Usenet newsgroup posts, consisting
of about 11K training and 7.5K test documents. RCV1 is a much larger dataset of
Reuters newswire articles, with about 794.4K training and 10K test documents.
We use the standard preprocessed versions of the datasets from
\citet{salakhutdinov2009replicated}, which have vocabularies of 2K and 10K words respectively.

We experimented with two simple document models, based  on the SBN and DARN
architectures. Both models had a single layer of latent variables and a
multinomial visible layer and can be seen as directed counterparts of the
Replicated Softmax model \citep{salakhutdinov2009replicated}.
We used the same training procedure as on MNIST with the exception
of the learning rates which were $3\times10^{-5}$ on 20 Newsgroups and
$10^{-3}$ on RCV1.

The established evaluation metric for such models is the perplexity per word,
computed as $\exp\left(-\frac{1}{N}\sum_n \frac{1}{L_n} \log P(x_n)\right)$,
where $N$ is the number of documents, $L_n$ is the length of document $n$, and
$P(x_n)$ the probability of the document under the model. As we cannot
compute $\log P(x_n)$, we use the variational lower bound in its place and
thus report an upper bound on perplexity.

The results for our models, along with ones for the Replicated Softmax and
DocNADE models from \citep{salakhutdinov2009replicated} and \citep{larochelle2012neural} respectively, are shown in
\tbl{tbl:docResults}. We can see that the SBN and fDARN models with 50 latent
variables perform well, producing better scores than LDA and Replicated Softmax
on both datasets. Their performance is also competitive with that of DocNADE on
20 Newsgroups. The score of 724 for fDARN with 50 latent variables on RCV1 is
already better than DocNADE's 742, the best published result on that dataset.
fDARN with 200 hidden units, however, performs even better, setting a new
record with 598.

\begin{table}[t]
\caption{Document modelling results.
``Dim'' is the number of latent variables in the model.
The third and the fourth columns report the estimated test set perplexity
on the 20 Newsgroups and Reuters RCV1 datasets respectively. }
\label{tbl:docResults}
\vskip 0.15in
\begin{center}
\begin{small}
\begin{sc}
\begin{tabular}{|l|r|r|r|}
\hline
Model           &  Dim & 20 News & Reuters \\
\hline
SBN             &  50  &  909   &   784 \\
fDARN           &  50  &  917   &   724 \\
fDARN           & 200  &        &   598 \\
\hline                    
LDA             &  50  &  1091  &  1437 \\
LDA             & 200  &  1058  &  1142 \\
RepSoftMax      &  50  &  953   &   988 \\
DocNade         &  50  &  896   &   742 \\
\hline                    

\hline
\end{tabular}
\end{sc}
\end{small}
\end{center}
\vskip -0.1in
\end{table}

\section{Discussion and future work}

We developed, NVIL, a new training method for intractable directed latent variable
models which is general and easy to apply to new models.
We showed that NVIL consistently outperforms the wake-sleep algorithm at
training sigmoid-belief-network-like models. Finally, we demonstrated the
potential of our approach by achieving state-of-the-art results on a sizable
dataset of documents (Reuters RCV1).

As the emphasis of this paper is on the training method, we applied it to some
of the simplest possible model and inference network architectures, which was
sufficient to obtain promising results.
We believe
that considerable performance gains can be made by using more expressive
architectures, such as those with nonlinearities between layers of stochastic
variables. Applying NVIL to models with continuous latent variables is another
promising direction since binary latent variables are not always appropriate.

We expect NVIL to be also applicable to training conditional latent variable models
for modelling the distribution of observations given some context, which would
require making the inference network take both the context and the observation
as input. This would make it an alternative to the importance-sampling training
method of \citet{tang2013learning} for conditional models with structured
high-dimensional outputs.

We hope that the generality and flexibility of our approach will make it easier
to apply powerful directed latent variable models to real-world problems.

\subsubsection*{Acknowledgements}
We thank Koray Kavukcuoglu, Volodymyr Mnih, and Nicolas Heess for their helpful
comments. We thank Ruslan Salakhutdinov for providing us with the preprocessed
document datasets.

% Override Natbib's bibliography heading
\renewcommand{\bibsection}{\section*{References}}

\bibliography{paper}
\bibliographystyle{icml2014}

\section*{A. Algorithm for computing NVIL gradients}

Algorithm~\ref{alg1} provides an outline of our implementation of NVIL gradient
computation for a minibatch of $n$ randomly chosen training
cases. The exponential smoothing factor $\alpha$ used for updating the
estimates of the mean $c$ and variance $v$ of the inference network learning signal was
set to 0.8 in our experiments.

\begin{algorithm}[th!]
\caption{Compute gradient estimates for the model and the inference network}
\label{alg1}
\begin{algorithmic}
    \STATE $\Delta\theta \leftarrow 0, \Delta\phi \leftarrow 0, \Delta\psi \leftarrow 0$
    \STATE $\varL \leftarrow 0$ \\
    \COMMENT{Compute the learning signal and the bound}
    \FOR {$i \leftarrow 1$ to $n$}
        \STATE $x_i \leftarrow $ random training case \\
        \COMMENT{Sample from the inference model}
        \STATE $h_i \sim Q_{\phi}(h_i|x_i)$ \\
        \COMMENT{Compute the unnormalized learning signal}
        \STATE $l_i \leftarrow \log P_{\theta}(x_i,h_i) - \log Q_{\phi}(h_i|x_i)$ \\
        \COMMENT{Add the case contribution to the bound}
        \STATE $\varL \leftarrow \varL + l_i$ \\
        \COMMENT{Subtract the input-dependent baseline}
        \STATE $l_i \leftarrow l_i - C_\psi(x_i)$
    \ENDFOR \\
    \COMMENT{Update the learning signal statistics}
    \STATE $c_b \leftarrow $ mean$(l_1, ..., l_n)$
    \STATE $v_b \leftarrow $ variance$(l_1, ..., l_n)$
    \STATE $c \leftarrow \alpha c + (1-\alpha)c_b$
    \STATE $v \leftarrow \alpha v + (1-\alpha)v_b$
    \FOR {$i \leftarrow 1$ to $n$}
        \STATE $l_i \leftarrow \frac{l_i - c}{\text{max}(1, \sqrt{v})}$ \\
        \COMMENT{Accumulate the model parameter gradient}
        \STATE $\Delta\theta \leftarrow \Delta\theta + \ddTheta \log P_{\theta}(x_i,h_i)$  \\
        \COMMENT{Accumulate the inference net gradient}
        \STATE $\Delta\phi \leftarrow \Delta\phi + l_i \ddPhi \log Q_{\phi}(h_i|x_i)$  \\
        \COMMENT{Accumulate the input-dependent baseline gradient}
        \STATE $\Delta\psi \leftarrow \Delta\psi + l_i \ddPsi C_{\psi}(x_i)$
    \ENDFOR 
\end{algorithmic}
\end{algorithm}

\section*{B. Inference network gradient derivation}
Differentiating the variational lower bound \wrt to the inference network parameters gives
\begin{align*}
    \ddPhi \varL(x) = & \ddPhi E_Q [\log P_{\theta}(x,h) - \log Q_{\phi}(h|x)] \\
                = & \ddPhi \sum_h Q_{\phi}(h|x) \log P_{\theta}(x,h) - \\
                  & \ddPhi \sum_h Q_{\phi}(h|x) \log Q_{\phi}(h|x) \\
                = & \sum_h \log P_{\theta}(x,h) \ddPhi Q_{\phi}(h|x) - \\
                  & \sum_h \left( \log Q_{\phi}(h|x)  + 1 \right) \ddPhi Q_{\phi}(h|x) \\
                = & \sum_h \left( \log P_{\theta}(x,h) - \log Q_{\phi}(h|x) \right) \ddPhi Q_{\phi}(h|x),
\end{align*}
where we used the fact that $\sum_h \ddPhi Q_{\phi}(h|x) = \ddPhi \sum_h Q_{\phi}(h|x) = \ddPhi 1 = 0$.
Using the identity $\ddPhi Q_{\phi}(h|x) =  Q_{\phi}(h|x) \ddPhi \log Q_{\phi}(h|x)$, then gives
\begin{align*}
    \ddPhi \varL(x) = & \sum_h \left( \log P_{\theta}(x,h) - \log Q_{\phi}(h|x) \right) \nonumber \\
                  & \times Q_{\phi}(h|x) \ddPhi \log Q_{\phi}(h|x) \nonumber \\
                = & E_Q \left [ \left(\log P_{\theta}(x,h) - \log Q_{\phi}(h|x)\right) \ddPhi \log Q_{\phi}(h|x)  \right].
\end{align*}

\end{document}